\documentclass{article}%
\usepackage{amsmath}
\usepackage{amsfonts}
\usepackage{amssymb}
\usepackage{graphicx}
\usepackage{algorithmic}
\usepackage{algorithm}%
\providecommand{\U}[1]{\protect\rule{.1in}{.1in}}

\begin{document}

\title{BENK: the Beran Estimator with Neural Kernels for Estimating the Heterogeneous
Treatment Effect}
\author{Stanislav R. Kirpichenko, Lev V. Utkin, Andrei V. Konstantinov\\Peter the Great St.Petersburg Polytechnic University\\St.Petersburg, Russia\\e-mail: kirpichenko.sr@gmail.com, lev.utkin@gmail.com, andrue.konst@gmail.com}
\date{}
\maketitle

\begin{abstract}
A method for estimating the conditional average treatment effect under
condition of censored time-to-event data called BENK (the Beran Estimator with
Neural Kernels) is proposed. The main idea behind the method is to apply the
Beran estimator for estimating the survival functions of controls and
treatments. Instead of typical kernel functions in the Beran estimator, it is
proposed to implement kernels in the form of neural networks of a specific
form called the neural kernels. The conditional average treatment effect is
estimated by using the survival functions as outcomes of the control and
treatment neural networks which consists of a set of neural kernels with
shared parameters. The neural kernels are more flexible and can accurately
model a complex location structure of feature vectors. Various numerical
simulation experiments illustrate BENK and compare it with the well-known
T-learner, S-learner and X-learner for several types of the control and
treatment outcome functions based on the Cox models, the random survival
forest and the Nadaraya-Watson regression with Gaussian kernels The code of
proposed algorithms implementing BENK is available in https://github.com/Stasychbr/BENK.


\end{abstract}

\section{Introduction}

Survival analysis is an important and fundamental tool for modelling
applications using time-to-event data \cite{Hosmer-Lemeshow-May-2008} which
can be encountered in medicine, reliability, safety, finance, etc. This is a
reason why many machine learning models have been developed to deal with
time-to-event data and to solve the corresponding problems in the framework of
survival analysis \cite{Katzman-etal-2018,Wang-Li-Reddy-2019,Zhao-Feng-2019}.
The crucial peculiarity of time-to-event data is that a training set consists
of censored and uncensored observations. When time-to-event exceeds the
duration of observation, we have a censored observation. When an event is
observed, i.e., time-to-event coincides with the duration of the observation,
we deal with an uncensored observation.

Many survival models are available to cover various cases of the time-to-event
probability distributions and their parameters \cite{Wang-Li-Reddy-2019}. One
of the important models is the Cox proportional hazards model \cite{Cox-1972}
which can be regarded as a regression semi-parametric model. There are also
many parametric and non-parametric models. If to consider the machine learning
survival models, then it is important to point out that, in contrast to other
machine learning models, their outcomes are functions, for example, survival
functions, hazard functions, cumulative hazard functions. For instance, the
well-known effective model called the random survival forest (RSFs)
\cite{Ishwaran-Kogalur-2007} predicts survival functions (SFs) or cumulative
hazard functions.

An important area of the survival model application is the problem of the
treatment effect estimation, which is often solved in the framework of machine
learning problems \cite{Lu-Sadiq-etal-2017,Shalit-etal-2017}. The treatment
effect shows how a treatment may be efficient depending on characteristics of
a patient. The problem is solved by dividing patients into two groups called
treatment and control such that patients from the different groups can be
compared. One of the popular measures of the efficient treatment used in
machine learning models is the average treatment effect (ATE)
\cite{Fan-Lv-Wang-2018}, which is estimated on the basis of observed data
about patients as the mean difference between outcomes of patients from the
treatment and control groups.

Due to the difference between characteristics of patients and between their
responses to a particular treatment, the treatment effect is measured by the
conditional average treatment effect (CATE) which is defined as the mean
difference between outcomes of patients from the treatment and control groups
conditional on a patient feature vector \cite{Wager-Athey-2015}. In fact, most
methods of the CATE estimation are based on constructing two regression models
for controls and treatments. However, two difficulties of the CATE estimation
can be met. The first one is that the treatment group is usually very small.
Therefore, many machine learning models cannot be accurately trained on the
small datasets. The second difficulty is fundamental. Each patient cannot be
simultaneously in the treatment and control groups, i.e., we either observe
the patient outcome under treatment or control, but never both
\cite{Kunzel-etal-2018a}. Nevertheless, to overcome the difficulties, many
methods for estimating CATE have been proposed and developed due to importance
of the problem in many areas
\cite{Acharki-etal-22,Alaa-Schaar-2018,Athey-Imbens-2016,Hatt-etal-22,Jiang-Qi-etal-21,Kunzel-etal-19,Wu-Yang-22,Zhang-Li-Liu-22}%
.

\begin{figure}
[ptb]
\begin{center}
\includegraphics[
height=2.2611in,
width=2.6938in
]%
{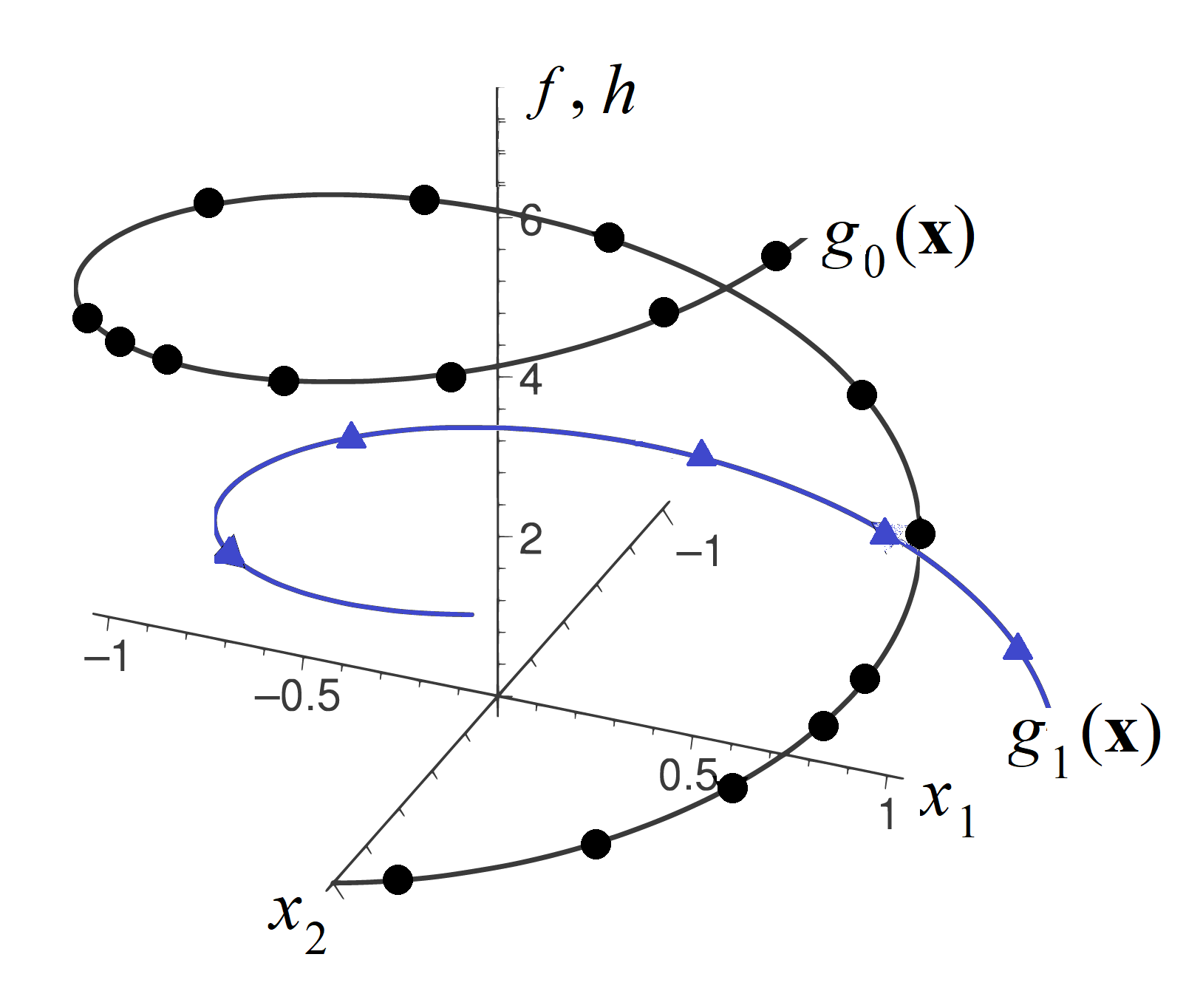}%
\caption{An example of the control $g_{0}(\mathbf{x})$ and treatment
$g_{1}(\mathbf{x})$ functions, which are unknown, and of the control (circle
markers) and treatment (triangle markers) data points, which are observed}%
\label{f:t_c_spirals}%
\end{center}
\end{figure}

One of the approaches for constructing regression models for controls and
treatments is the application of the Nadaraya-Watson kernel regression
\cite{Nadaraya-1964,Watson-1964}, which uses standard kernel functions, for
instance, the Gaussian, uniform, or Epanechnikov kernels. In order to avoid
selecting a standard kernel, Konstantinov et al. \cite{Konstantinov-etal-22}
proposed to implement kernels and the whole Nadaraya-Watson kernel regression
by using a set of identical neural subnetworks with shared parameters with a
specific way of the network training. The corresponding method called TNW-CATE
(the Trainable Nadaraya-Watson regression for CATE) is based on an important
assumption that domains of the feature vectors from the treatment and control
groups are similar. Indeed, we often treat patients after being in the control
group, i.e., it is assumed that treated patients came to the treatment group
from the control group. The neural kernels (kernels implemented as the neural
network) are more flexible, and they can accurately model a complex location
structure of feature vectors, for instance, when feature vectors from the
control and treatment group are located on the spiral as it is shown in Fig.
\ref{f:t_c_spirals} where small triangular and circle markers correspond to
the treatment and control groups, respectively. This is another important
peculiarity of TNW-CATE. Results provided in \cite{Konstantinov-etal-22}
illustrated outperformance of TNW-CATE in comparison with other methods when
the treatment group is very small and the feature vectors have complex structure.

Following the ideas of TNW-CATE, we propose the CATE estimation method, called
BENK (the Beran Estimator with Neural Kernels), dealing with censored time-to
event data. The main idea behind the method is to apply the Beran estimator
\cite{Beran-81} for training the neural kernels. The Beran estimator allows us
to get SFs conditional on the feature vectors, which can be regarded as
outcomes of regression survival models for the treatment and control group.
The Beran estimator depends on a kernel which is used for estimating. We again
propose to implement the kernels by means of neural subnetworks and estimate
CATE by using the obtained SFs.%

Various numerical experiments illustrate BENK and its peculiarities. They also
show that BENK outperforms many well-known meta-models: the T-learner, the
S-learner, the X-learner for several control and treatment output functions
based on the Cox models, the RSF and the Nadaraya-Watson regression with
Gaussian kernels.

The code of the proposed algorithms can be found in https://github.com/Stasychbr/BENK.

The paper is organized as follows. Section 2 is a review of the existing CATE
estimation models, including CATE estimation survival models\textbf{,} the
Nadaraya-Watson regression models and general survival models. A formal
statement of the CATE estimation problem is provided in Section 3. The CATE
estimation problem in case of censored data is stated in Section 4. The Beran
estimator is considered in Section 5. A description of BENK is provided in
Section 6. Numerical experiments illustrating BENK and comparing it with other
models can be found in Section 7. Concluding remarks are provided in Section 8.

\section{Related work}

\textbf{Estimating CATE}. One of the important approaches to implement the
personalized medicine \cite{Powers-etal-2017} is the treatment effect
estimation. As a result, many interesting machine learning models have been
developed and implemented to estimate CATE. First, we have to point out an
approach which uses the Lasso model for estimating CATE
\cite{Jeng-Lu-Peng-2018}. The SVM was also applied to solving the problem
\cite{Zhou-Mayer-Hamblett-etal-2017}. A unified framework for constructing
fast tree-growing procedures for solving the CATE problem was provided in
\cite{Athey-Tibshirani-Wager-2016,Athey-Tibshirani-Wager-2018}. An interesting
model for computing CATE was proposed in \cite{Athey-Imbens-2016} where the
training set is splitting into two subsets such that the first one is used to
construct the partition of the data into subpopulations that differ in the
magnitude of their treatment effect, and the second subset is used to estimate
treatment effects for each subpopulation. The CATE detection problem was
considered as a false positive rate control problem in
\cite{Xie-Chen-Shi-2018}. Alaa and Schaar \cite{Alaa-Schaar-2018} proposed
algorithms for estimating CATE in the context of Bayesian nonparametric
inference. Bayesian additive regression trees, a causal forest, and a causal
boosting models were compared under condition of binary outcomes in
\cite{Wendling-etal-2018}. An orthogonal random forest as an algorithm that
combines orthogonalization with generalized random forests for solving the
CATE estimation problem was proposed in \cite{Oprescu-Syrgkanis-Wu-2018}.
McFowland et al. \cite{McFowland-etal-2018} estimated CATE by using the
anomaly detection model. A set of meta-algorithms or meta-learners, including
the T-learner \cite{Kunzel-etal-2018}, the S-learner \cite{Kunzel-etal-2018},
the O-learner \cite{Wang-etal-2016}, the X-learner \cite{Kunzel-etal-2018}
were studied in \cite{Kunzel-etal-2018}. Many other models related to the CATE
estimation problem are studied in
\cite{Chen-Liu-2018,Grimmer-etal-017,Levy-2018,Powers-etal-2017,Yao-Lo-Nir-etal-22}%
.

The aforementioned models are constructed by using the machine learning
methods different from neural networks. However, neural networks became a
basis for developing many interesting and efficient models
\cite{Bica-etal-20,Chen-Dong-Lu-etal-19,Curth-Schaar-21a,Du-Fan-etal-21,Nair-etal-22,Nie-etal-21,Parbhoo-etal-21,Qin-Wang-Zhou-21,Schwab-etal-2020}%
.

The next generation of models solving the CATE estimation problem is based on
architectures of transformers \cite{Chaudhari-etal-2019} with the attention
operations
\cite{Guo-Zheng-etal-21,Melnychuk-etal-22,Zhang-Zhang-etal-22,Zhang-Zhang-etal-22a}%
. The transfer learning technique was successfully applied to the CATE
estimation in
\cite{Aoki-Ester-22,Guo-Wang-etal-21,Kunzel-etal-18,Kunzel-etal-2018a,Zhou-Yau-Xu-etal-22}%
. Ideas of using the Nadaraya-Watson kernel regression in the CATE estimation
were studied in \cite{Imbens-04,Park-Shalit-etal-21}. These ideas can lead to
the best results under condition of large numbers of examples in the treatment
and control groups. At the same time, a small amount of training data may lead
to overfitting and unsatisfactory results. Therefore, a problem of overcoming
this possible limitation motivated to introduce a neural network of a special
architecture, which implements the trainable kernels in the Nadaraya-Watson
regression \cite{Konstantinov-etal-22}.

\textbf{Machine learning models in survival analysis}. Importance of the
survival analysis applications can be regarded as one of the reasons for
developing many machine learning methods dealing with censored and
time-to-event data. A comprehensive review of machine learning survival models
is presented in \cite{Wang-Li-Reddy-2019}. A large part of models uses the Cox
model which can be viewed as a simple and applicable survival model which
establish a relationship between covariates and outcomes. Various extensions
of the Cox model have been proposed. They can be conditionally divided into
two groups. The first group remains the linear relationship of covariates and
includes various modifications of the Lasso models
\cite{Kaneko-etal-2015,Witten-Tibshirani-2010}. The second group of models
relaxes the linear relationship assumption accepted in the Cox model
\cite{Faraggi-Simon-1995,Widodo-Yang-2011}.

Many survival models are based on using the RSFs which can be regarded as
powerful tools especially when models learn on tabular data
\cite{Ibrahim-etal-2008,Mogensen-etal-2012,Wang-Zhou-2017,Wright-etal-2017}.
At the same time, there are many survival models based on neural networks
\cite{Haarburger-etal-2018,Katzman-etal-2018,Ranganath-etal-2016,Zhu-Yao-Huang-2016}%
.

\textbf{The Nadaraya-Watson regression in machine learning}. The
Nadaraya-Watson regression can be viewed as an effective tool for solving many
machine learning tasks
\cite{Conn-Li-19,Hanafusa-Okadome-20,Konstantinov-Utkin-22d,Liu-Huang-etal-21a,Liu-Min-etal-21,Xiao-Xiang-etal-19}%
. In particular, the Nadaraya-Watson regression was used as a trainable
convolution neural network layer in \cite{Szczotka-etal-20}. An interesting
approach to simplify the Nadaraya-Watson estimator by approximating the kernel
functions produced from data points which are located in the neighborhood of
input values was presented by Ito et al. \cite{Ito-etal-20}. Noh et al.
\cite{Noh-etal-17} considered how the Nadaraya-Watson kernel regression can be
applied to solve a metric learning model. An application of the
Nadaraya-Watson regression to the local explanation, in particular, to the
method SHAP, was proposed by Ghalebikesabi et al. \cite{Ghalebikesabi-etal-21}%
. The most important application of the Nadaraya-Watson regression is the
attention mechanism explanation, i.e., it can be regarded as a way for
explaining ideas of the attention mechanism from the statistics point of view
\cite{Chaudhari-etal-2019,Zhang2021dive}.

\textbf{Estimating CATE with censored data.} Censored data can be regarded as
an important type especially for estimating the treatment effect because many
applications are characterized by the time-to-event data as outcomes. This
peculiarity is a reason for developing many CATE models dealing with censored
data in the framework of survival analysis. In particular, a modification of
the survival causal tree method for estimating the CATE based on censored
observational data was proposed in \cite{Zhang-Le-etal-2017}. An approach
combining a treatment-specific semi-parametric Cox loss with a
treatment-balanced deep neural network was studied in \cite{Schrod-etal-22}.
Nagpal et al. \cite{Nagpal-etal-22} presented a latent variable approach to
model CATE under assunption that an individual can belong to one of latent
clusters with distinct response characteristics. Causal survival forests was
introduced in \cite{Cui-etal-20}, which can be used to estimate CATE under
condition of right-censored data. The problem of estimating CATE with focusing
on learning (discrete-time) treatment-specific conditional hazard functions
was studied in \cite{Curth-etal-21}. A three-stage modular design for
estimating CATE in the framework of survival analysis was proposed in
\cite{Zhu-Gallego-20}. A comprehensive simulation study presenting a wide
range of settings describing CATE taking into account the covariate overlap
was carried out in \cite{Hu-Ji-Li-21}. Doubly robust estimation equations were
derived in \cite{Ozenne-etal-2020} where estimators for the nuisance
parameters based on working regression models for the outcome, censoring, and
treatment distribution conditional on auxiliary baseline covariates are
implemented. Rytgaard et al. \cite{Rytgaard-etal-2021} presented a
data-adaptive estimation procedure for estimation of CATE in a time-to-event
setting based on generalized random forests. The authors proposed a two-step
procedure for estimation, applying inverse probability weighting to construct
time-point specific weighted outcomes as input for the forest. A unified
framework for counterfactual inference applicable to survival outcomes and
formulating a nonparametric hazard ratio metric for evaluating CATE were
proposed in \cite{Chapfuwa-etal-21}.

In spite of many works and results devoted to estimating CATE with censored
data, they are mainly based on assumptions of a large amount of examples in
the treatment group. Moreover, there are no results implementing the
Nadaraya-Watson regression by means of neural networks.

\section{CATE estimation problem statement}

According to the CATE estimation problem, all patients are divided into two
groups: control and treatment. Let the control group be the set $\mathcal{C}%
=\{(\mathbf{x}_{1},f_{1}),...,(\mathbf{x}_{c},f_{c})\}$ of $c$ patients such
that the $i$-th patient is characterized by the feature vector $\mathbf{x}%
_{i}=(x_{i1},...,x_{id})\in\mathbb{R}^{d}$ and the $i$-th observed outcome
$f_{i}\in\mathbb{R}$ (time to event, temperature, the blood pressure, etc.).
It is also supposed that the treatment group is the set $\mathcal{T}%
=\{(\mathbf{y}_{1},h_{1}),...,(\mathbf{y}_{t},h_{t})\}$ of $t$ patients such
that the $i$-th patient is characterized by the feature vector $\mathbf{y}%
_{i}=(y_{i1},...,y_{id})\in\mathbb{R}^{d}$ and the $i$-th observed outcome
$h_{i}\in\mathbb{R}$. The indicator of a group for the $i$-th patient is
denoted as $T_{i}\in\{0,1\}$, where $T_{i}=0$ ($T_{i}=1$) corresponds to the
control (treatment) group.

We use different notations $\mathbf{x}_{i}$ and $\mathbf{y}_{i}$ for controls
and treatments in order to avoid additional indices. However, we will use the
vector $\mathbf{z}\in\mathbb{R}^{d}$ instead of $\mathbf{x}$ and $\mathbf{y}$
when estimate the CATE.

Suppose that the potential outcomes of patients from the control and treatment
groups are $F$ and $H$, respectively. The treatment effect for a new patient
with the feature vector $\mathbf{z}$ is estimated by the individual treatment
effect defined as $H-F$. The fundamental problem of computing CATE is that
only one of outcomes $f$ or $h$ for each patient can be observed. An important
assumption of unconfoundedness \cite{Rosenbaum-Rubin-1983} is used to allow
the untreated patients to be used to construct an unbiased counterfactual for
the treatment group \cite{Imbens-2004}. According to the assumption, potential
outcomes are characteristics of a patient before the patient is assigned to a
treatment condition, or formally the treatment assignment $T$ is independent
of the potential outcomes for $F$ and $H$ conditional on the feature vector
$\mathbf{z}$, respectively, which can be written as
\begin{equation}
T\perp\{F,H\}|\mathbf{z}.
\end{equation}

The second assumption, called the overlap assumption, regards the joint
distribution of treatments and covariates. This assumption claims that a
positive probability of being both treated and untreated for each value of
$\mathbf{z}$ exists. It is of the form:
\begin{equation}
0<\Pr\{T=1|\mathbf{z}\}<1.
\end{equation}

Let $\mathbf{Z}$ be the random feature vector from $\mathbb{R}^{d}$. The
treatment effect is estimated by means of CATE which is defined as the
expected difference between two potential outcomes as follows
\cite{Rubin-2005}:
\begin{equation}
\tau(\mathbf{z})=\mathbb{E}\left[  H-F|\mathbf{Z}=\mathbf{z}\right]  .
\end{equation}

By using the above assumptions, CATE\ can be rewritten as:
\begin{equation}
\tau(\mathbf{z})=\mathbb{E}\left[  H|\mathbf{Z}=\mathbf{z}\right]
-\mathbb{E}\left[  F|\mathbf{Z}=\mathbf{z}\right]  .
\end{equation}

The motivation behind unconfoundedness is that nearby observations in the
feature space can be treated as having come from a randomized experiment
\cite{Wager-Athey-2017}.

Suppose that functions $g_{0}(\mathbf{z})$ and $g_{1}(\mathbf{z})$ express
outcomes of the control and treatment patients, respectively. Then they can be
written as follows:%
\begin{equation}
f=g_{0}(\mathbf{z})+\varepsilon,~h=g_{1}(\mathbf{z})+\varepsilon,
\end{equation}
where $\varepsilon$ is a noise governed by the normal distribution with the
zero expectation.

The above implies that CATE can be estimated as:
\begin{equation}
\tau(\mathbf{z})=g_{1}(\mathbf{z})-g_{0}(\mathbf{z}). \label{HTE_concat_20}%
\end{equation}

An example illustrating controls (circle markers), treatments (triangle
markers) and the corresponding unknown function $g_{0}$ and $g_{1}$ are shown
Fig. \ref{f:t_c_spirals}.

\section{CATE with censored data}

Before considering the CATE estimation problem with censored data, we
introduce basic statements of survival analysis. Let us define the training
set $D_{0}$ which consists of $c$ triplets $(\mathbf{x}_{i},\delta_{i},f_{i}%
)$, $i=1,...,c$, where $\mathbf{x}_{i}^{\mathrm{T}}=(x_{i1},...,x_{id})$ is
the feature vector characterizing the $i$-th patient from the control group;
$f_{i}$ is the time to the event concerning the $i$-th control patient;
$\delta_{i}\in\{0,1\}$ is the indicator of censored or uncensored
observations. If $\delta_{i}=1$, then the event of interest is observed (the
uncensored observation). If $\delta_{i}=0$, then we have the censored
observation. Only the right censoring is considered when the observed survival
time is less than or equal to the true survival time. Many applications of
survival analysis deal with the right censored observations
\cite{Wang-Li-Reddy-2019}. The main goal of the survival machine learning
modelling is to use set $D_{0}$ to estimate probabilistic characteristics of
time $F$ to the event of interest for a new patient with the feature vector
$\mathbf{z}$.

In the same way, we define the training set $D_{1}$ which consists of $d$
triplets $(\mathbf{y}_{i},\gamma_{i},h_{i})$, $i=1,...,s$, where
$\mathbf{y}_{i}^{\mathrm{T}}=(y_{i1},...,y_{id})$ is the feature vector
characterizing the $i$-th patient from the treatment group; $h_{i}$ is the
time to the event concerning the $i$-th treatment patient; $\gamma_{i}%
\in\{0,1\}$ is the indicator of censoring.

The survival function (SF), denoted $S(t|\mathbf{z})$, can be regarded as an
important concept in survival analysis. It represents the probability of
surviving of a patient with the feature vector $\mathbf{z}$ up to time $t$
that is $S(t|\mathbf{z})=\Pr\{T>t|\mathbf{z}\}$. The hazard function, denoted
$\lambda(t|\mathbf{z})$, can be viewed as another concept in survival
analysis. It is defined as the rate of event at time $t$ given that no event
occurred before time $t$. It is expressed through the SF as follows:
\begin{equation}
\lambda(t|\mathbf{z})=-\frac{\mathrm{d}}{\mathrm{d}t}\ln S(t|\mathbf{z}).
\end{equation}

The integral of the hazard function, denoted $H(t|\mathbf{x})$, is called the
cumulative hazard function and can be interpreted as the probability of an
event at time $t$ given survival until time $t$, i.e.,
\begin{equation}
\Lambda(t|\mathbf{z})=\int_{-\infty}^{t}\lambda(r|\mathbf{z})dr.
\end{equation}

It is expressed through the SF as follows:%
\begin{equation}
\Lambda(t|\mathbf{z})=-\ln\left(  S(t|\mathbf{z})\right)  .
\end{equation}

The above functions for controls and treatments will be written with indices
$0$ and $1$, respectively, for example, $S_{0}(t|\mathbf{z})=\Pr
\{F>t|\mathbf{z}\}$ and $S_{1}(t|\mathbf{z})=\Pr\{H>t|\mathbf{z}\}$.

In order to compare survival models the Harrell's concordance index or the
C-index \cite{Harrell-etal-1982} is usually used. The C-index measures the
probability that, in a randomly selected pair of examples, the example that
fails first had a worst predicted outcome. It is calculated as the ratio of
the number of pairs correctly ordered by the model to the total number of
admissible pairs. A pair is not admissible if the events are both
right-censored or if the earliest time in the pair is censored. If the C-index
is equal to 1, then The corresponding survival model is supposed to be perfect
when the C-index is 1. The case when the C-index is 0.5 says that the survival
model is the same as random guessing. The case when the C-index is less than
0.5 says that the corresponding model is worse than random guessing.

In contrast to the standard CATE estimation problem statement given in the
previous section, the CATE estimation problem with censored data has another
statement which is caused by the fact that outcomes in survival analysis are
random times to event of interest having some conditional probability
distribution. In other words, predictions corresponding to a patient
characterizing by vector $\mathbf{z}$ in survival analysis provided by a
survival machine learning model are represented in the form of functions of
time, for example, in the form of SF $S(t|\mathbf{z})$. This implies that
CATE\ $\tau(\mathbf{x})$ should be reformulated taking into account the above
peculiarity. It is assumed that SFs as well as hazard functions for control
and treatment patients estimated by using datasets $D_{0}$ and $D_{1}$ will
have indices $0$ and $1$, respectively.

The following definitions of CATE in the case of censored data can be found in
\cite{Chapfuwa-etal-20,Trinquart-etal-16,Zhao-Tian-etal-12}:

\begin{enumerate}
\item Difference in expected lifetimes: $\tau(\mathbf{z})=$ $\int_{0}%
^{t_{\max}}\left(  S_{1}(t|\mathbf{z})-S_{0}(t|\mathbf{z})\right)
\mathrm{d}t=\mathbb{E}\left\{  T_{1}-T_{0}|X=\mathbf{z}\right\}  $.

\item Difference in SFs: $\tau(t,\mathbf{z})=$ $S_{1}(t|\mathbf{z}%
)-S_{0}(t|\mathbf{z})$.

\item Hazard ratio: $\tau(t,\mathbf{z})=\lambda_{1}(t|\mathbf{z})/\lambda
_{0}(t|\mathbf{z})$.
\end{enumerate}

We will use the first integral definition of CATE. Let $0=t_{0}<t_{1}%
<...<t_{n}$ be the distinct times to event of interest which are obtained from
the set $\{f_{1},...,f_{c}\}\cup\{h_{1},...,h_{s}\}$. The SF provided by a
survival machine learning model is a step function, i.e., it can be
represented as $S(t|\mathbf{z})=\sum_{j=1}^{n}S^{(j)}(\mathbf{z})\cdot\chi
_{j}(t)$, where $\chi_{j}(t)$ is the indicator function taking value 1 if
$t\in\lbrack t_{j-1},t_{j}]$; $S^{(j)}(\mathbf{z})$ is the value of the SF in
interval $[t_{j-1},t_{j}]$. Hence, there holds
\begin{align}
\tau(\mathbf{z}) &  =\int_{0}^{t_{\max}}\left(  S_{1}(t|\mathbf{z}%
)-S_{0}(t|\mathbf{z})\right)  \mathrm{d}t\nonumber\\
&  =\sum_{j=1}^{n}\left(  S_{1}^{(j)}(\mathbf{z})-S_{0}^{(j)}(\mathbf{z}%
)\right)  (t_{j}-t_{j-1}).\label{CATE_20}%
\end{align}

\section{ Nonparametric estimation of survival functions and CATE}

The idea to use the Nadaraya-Watson regression for estimating SFs and other
concepts of survival analysis has been proposed by several authors
\cite{Bobrowski-etal-15,Gneyou-14,Pelaez-Cao-Vilar-22,Selingerova-etal-21,Tutz-Pritscher-96}%
. One of the interesting estimators is the Beran estimator \cite{Beran-81} of
the SF which is defined as follows:
\begin{equation}
S(t|\mathbf{x})=\prod\nolimits_{f_{i}\leq t}\left\{  1-\frac{W(\mathbf{x}%
,\mathbf{x}_{i})}{1-\sum_{j=1}^{i-1}W(\mathbf{x},\mathbf{x}_{j})}\right\}
^{\delta_{i}}, \label{Beran_est}%
\end{equation}
where $W(\mathbf{x},\mathbf{x}_{i})$ are the Nadaraya-Watson weights defined
as
\begin{equation}
W(\mathbf{x},\mathbf{x}_{i})=\frac{K(\mathbf{x},\mathbf{x}_{i})}{\sum
_{j=1}^{n}K(\mathbf{x},\mathbf{x}_{j})}.
\end{equation}

The above expression is given for the controls. The same estimator can be
written for treatments, but $\mathbf{x}$, $\delta_{i}$, $f_{i}$ is replaced
with $\mathbf{y}$, $\gamma_{i}$, $h_{i}$, respectively.

The Beran estimator can be regarded as a generalization of the Kaplan-Meier
estimator because it is reduced to the Kaplan-Meier estimator if the
Nadaraya-Watson weights take values $W(\mathbf{x},\mathbf{x}_{i})=1/n$. It is
also interesting to note that the product in (\ref{Beran_est}) takes into
account only uncensored observations whereas the weights are normalized by
using uncensored as well as censored observations.

By using (\ref{Beran_est}) and (\ref{CATE_20}), we can construct a neural
network which is trained to implement the weights $W(\mathbf{z},\mathbf{x}%
_{i})$, $W(\mathbf{z},\mathbf{y}_{i})$ and to estimate SFs $S_{1}%
(t|\mathbf{z})$ and $S_{0}(t|\mathbf{z})$ for computing $\tau(\mathbf{z})$.

\section{Neural network for estimating CATE}

Let us consider how the Beran estimator with neural kernels can be implemented
by means of a neural network of a special type. Our first aim is to implement
kernels $K(\mathbf{x},\mathbf{x}_{i})$ by means of a neural subnetwork, which
is called the neural kernel and is a part of the whole network implementing
the Beran estimator. The second aim is to learn this network on the control
data. Having the trained kernel, we can apply it to compute the conditional
survival function for controls as well as for treatments because the kernels
in (\ref{Beran_est}) directly do not depend on times to events $f_{i}$ or
$h_{i}$. However, in order to train the kernel, we have to train the whole
network because the loss function is defined through SF $S_{0}(t|\mathbf{x})$
which represents the probability of surviving of a control patient up to time
$t$, and which is estimated by means of the Beran estimator. This implies that
the whole network contains blocks of neural kernels for computing kernels
$K(\mathbf{x},\mathbf{x}_{i})$, normalization for computing the
Nadaraya-Watson weights $W(\mathbf{x},\mathbf{x}_{i})$ and the Beran estimator
in accordance with (\ref{Beran_est}). In order to realize a training procedure
of the network, we randomly select a part ($n$ examples) from all control
training examples and form a single specific example from $n$ selected ones.
This random selection is repeated $N$ times to have $N$ examples for training.

Having the trained neural kernel, it can be successfully used for computing SF
$S_{0}(t|\mathbf{z})$ of controls and SF $S_{1}(t|\mathbf{z})$ of treatments
for arbitrary vectors of features $\mathbf{z}$ again applying the Beran estimator.%

\begin{figure}
[ptb]
\begin{center}
\includegraphics[
height=3.4221in,
width=5.5582in
]%
{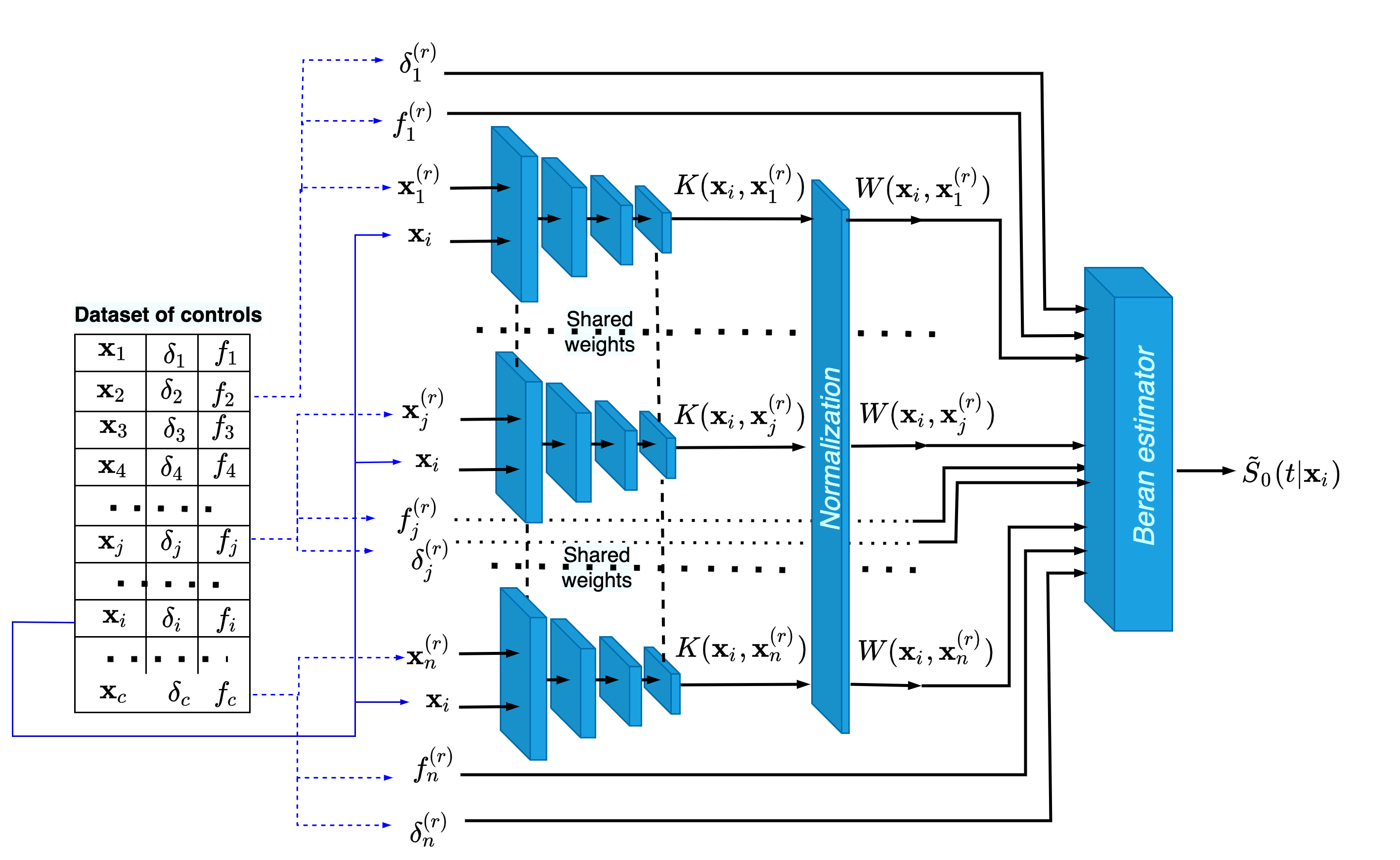}%
\caption{The neural network training on examples $\mathbf{a}_{i}^{(r)}$
composed of controls for producing the Beran estimator in the form of SF
$\tilde{S}_{0}(t|\mathbf{x}_{i})$}%
\label{f:train_control}%
\end{center}
\end{figure}

Let us consider the training algorithm in detail. First, we return to the set
of $c$ controls $\mathcal{C}=\{(\mathbf{x}_{i},\delta_{i},f_{i}%
),\ i=1,...,c\}$. For every $i$ from set $\{1,...,c\}$, we construct $N$
subsets $\mathcal{C}_{i}^{(r)}$, $r=1,...,N$, having $n$ examples randomly
selected from $\mathcal{C}\backslash(\mathbf{x}_{i},\delta_{i},f_{i})$, which
have indices from the index set $\mathcal{I}^{(r)}$, i.e., the subsets
$\mathcal{C}_{i}^{(r)}$ is of the form:
\begin{equation}
\mathcal{C}_{i}^{(r)}=\{(\mathbf{x}_{k}^{(r)},\delta_{k}^{(r)},f_{k}%
^{(r)}),\ k\in\mathcal{I}^{(r)}\},\ r=1,...,N.
\end{equation}

Here $N$ and $n$ can be regarded as tuning hyperparameters. Upper index $r$
indicates that the $r$-th example $(\mathbf{x}_{k}^{(r)},\delta_{k}%
^{(r)},f_{k}^{(r)})$ is randomly taken from $\mathcal{C}\backslash
(\mathbf{x}_{i},\delta_{i},f_{i})$, i.e., there is an example $(\mathbf{x}%
_{j},\delta_{j},f_{j})$ from $\mathcal{C}$ such that $\mathbf{x}_{k}%
^{(r)}=\mathbf{x}_{j}$, $\delta_{k}^{(r)}=\delta_{j}$, $f_{k}^{(r)}=f_{j}$.
Each subsets $\mathcal{C}_{r}^{(i)}$ jointly with $(\mathbf{x}_{i},\delta
_{i},f_{i})$ forms a training example $\mathbf{a}_{i}^{(r)}$ for the control
network as follows:
\begin{equation}
\mathbf{a}_{i}^{(r)}=\left(  \mathcal{C}_{i}^{(r)},\mathbf{x}_{i},\delta
_{i},f_{i}\right)  ,\ i=1,...,c,\ r=1,...,N. \label{THTE_42}%
\end{equation}

The number of possible examples $\mathbf{a}_{i}^{(r)}$ is $c\cdot N$, and
these examples will be used for training the neural network whose output is
the estimate of SF $\tilde{S}_{0}(t|\mathbf{x}_{i})$.

The architecture of the neural network consisting of $n$ subnetworks which
implement the neural kernels is shown in Fig. \ref{f:train_control}. Examples
$\mathbf{a}_{i}^{(r)}$ produced from the dataset of controls are fed to the
whole neural network such that each pair $(\mathbf{x}_{i},\mathbf{x}_{k}%
^{(r)})$, $k\in\mathcal{I}^{(r)}$, is fed to each subnetwork which implements
the kernel function. The output of each subnetwork is kernel $K(\mathbf{x}%
_{i},\mathbf{x}_{k}^{(r)})$. All subnetworks are identical and have shared
weights. After normalizing the kernels, we get $n$ weights $W(\mathbf{x}%
_{i},\mathbf{x}_{k}^{(r)})$ which are used to estimate SFs by means of the
Beran estimator in (\ref{Beran_est}). The block of the whole neural network
implementing the Beran estimator uses all weights $W(\mathbf{x}_{i}%
,\mathbf{x}_{k}^{(r)})$, $k\in\mathcal{I}^{(r)}$, and the corresponding values
$\delta_{k}^{(r)}$ and $f_{k}^{(r)}$, $k\in\mathcal{I}^{(r)}$. As a result, we
get SF $\tilde{S}_{0}(t|\mathbf{x}_{i})$. In the same way, we compute SFs
$\tilde{S}_{0}(t|\mathbf{x}_{k})$ for all $k=1,...,c$. These functions are the
basis for training. In fact, the normalization block and the block
implementing the Beran estimator are not a real part of the neural network and
they are not trained. They need to compute SFs and the corresponding loss
functions. However, we introduce these blocks into the presented architecture
in order to show that the loss function is determined through the Beran estimator.

According to (\ref{CATE_20}), expected lifetimes are used to compute CATE
$\tau(\mathbf{z})$. Therefore, the whole network is trained by means of the
following loss function:%
\begin{equation}
L=\frac{1}{c^{\ast}\cdot N}\sum_{i\in\mathcal{C}^{\ast}}\sum_{k=1}^{N}\left(
\tilde{E}_{k}^{(i)}-f_{k}^{(i)}\right)  ^{2}. \label{BENK_40}%
\end{equation}

Here $\mathcal{C}^{\ast}$ is a subset of $\mathcal{C}$, which contains only
uncensored examples from $\mathcal{C}$; $c^{\ast}$ is the number of elements
in $\mathcal{C}^{\ast}$; $f_{k}^{(i)}$ is the time to event of the $k$-th
example from the set $\mathcal{C}^{\ast}\backslash(\mathbf{x}_{i},\delta
_{i},f_{i})$; $\tilde{E}_{k}^{(i)}$ is the expected lifetime computed through
SF $\tilde{S}_{0}(t|\mathbf{x}_{k})$ by integrating the SF:
\begin{equation}
\tilde{E}_{k}^{(i)}=\sum_{j=1}^{n}(f_{j}^{(i)}-f_{j-1}^{(i)})\tilde{S}%
_{0}(f_{j}^{(i)}|\mathbf{x}_{k}).
\end{equation}

The sum in (\ref{BENK_40}) is taken over uncensored examples from
$\mathcal{C}$. However, the Beran estimator uses all examples.

It is important to point out that our aim is to train subnetworks with shared
training parameters, which are the neural kernels. By having the trained
neural kernels, we can use them to compute kernels $K(\mathbf{z}%
,\mathbf{x}_{i})$ and $K(\mathbf{z},\mathbf{y}_{i})$ and then to compute
estimates of SFs $\tilde{S}_{0}(t|\mathbf{z})$ and $\tilde{S}_{1}%
(t|\mathbf{z})$ for controls and treatments, respectively, i.e., we realize
the idea to transfer tasks from the control group to the treatment group. Let
$t_{1}^{(0)}<t_{2}^{(0)}<...<t_{c}^{(0)}$ and $t_{1}^{(1)}<t_{2}%
^{(1)}<...<t_{s}^{(1)}$ be the ordered time moments corresponding to times
$f_{1},...,f_{c}$ and $h_{1},...,h_{s}$, respectively. Then CATE
$\tau(\mathbf{z})$ can be computed through SFs $S_{1}(t|\mathbf{z})$ and
$S_{0}(t|\mathbf{z})$ again by using the Beran estimators with the trained
neural kernels, i.e., there holds in accordance with (\ref{CATE_20}):
\begin{equation}
\tau(\mathbf{z})=\sum_{j=1}^{s}(t_{j}^{(1)}-t_{j-1}^{(1)})\tilde{S}_{1}%
^{(j)}(\mathbf{z})-\sum_{k=1}^{c}(t_{k}^{(0)}-t_{k-1}^{(0)})\tilde{S}%
_{0}^{(k)}(\mathbf{z}),\label{BENK_41}%
\end{equation}
where $\tilde{S}_{1}^{(j)}(\mathbf{z})$ is the estimation of the SF of
treatments in interval $[t_{j-1}^{(1)},t_{j}^{(1)})$; $\tilde{S}_{0}%
^{(k)}(\mathbf{z})$ is the estimation of SF of controls in interval
$[t_{k-1}^{(0)},t_{k}^{(0)})$; it is assumed $t_{0}^{(0)}=t_{0}^{(1)}=0$. 

\begin{figure}
[ptb]
\begin{center}
\includegraphics[
height=2.9928in,
width=3.218in
]%
{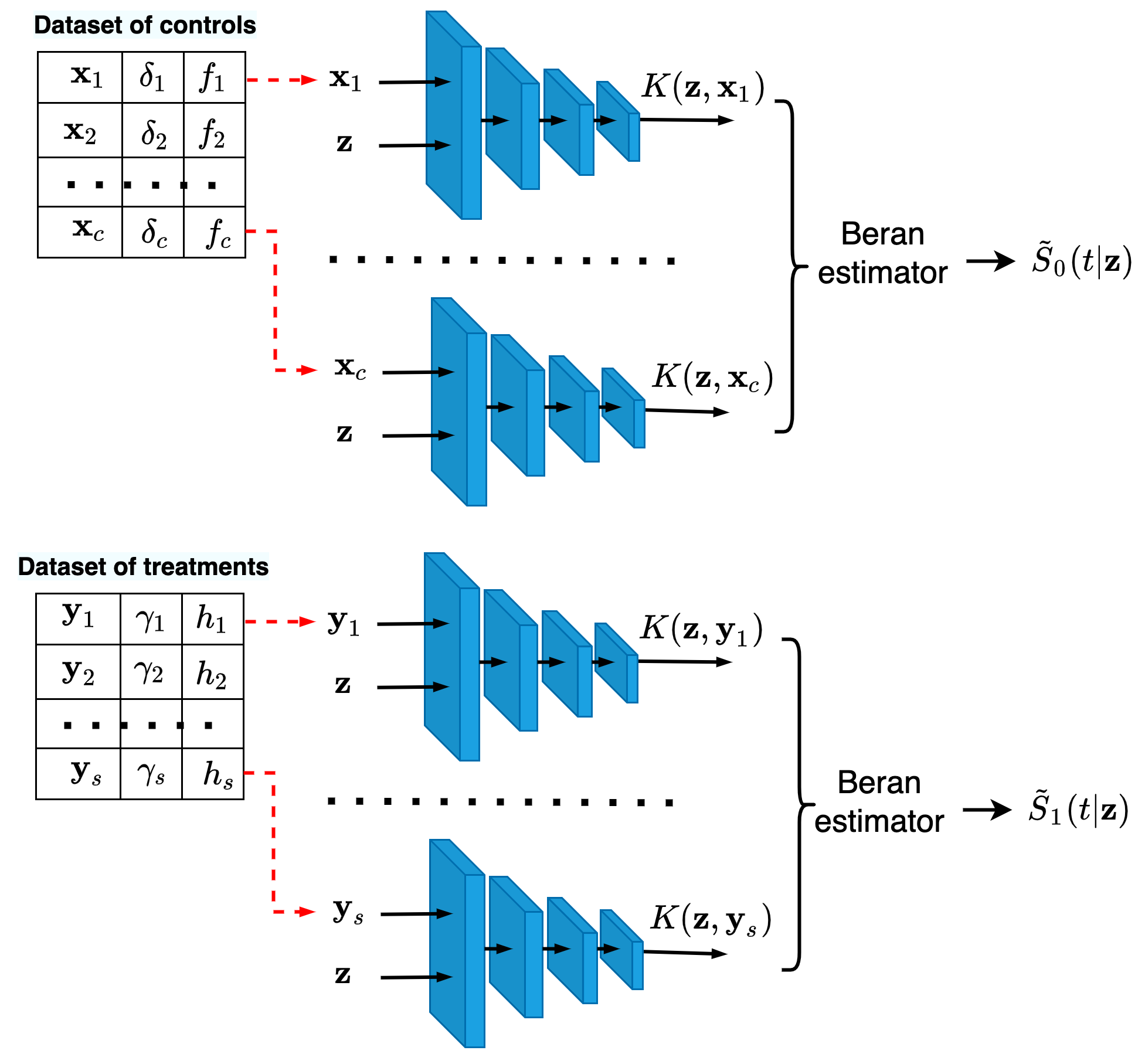}%
\caption{Neural networks consisting of the $c$ and $s$ trained neural kernels
predicting new values of kernels $K(\mathbf{z},\mathbf{x}_{i})$ and
$K(\mathbf{z},\mathbf{y}_{i})$ corresponding to controls and treatments for
computing estimates of $S_{1}(t|\mathbf{z})$ and $S_{0}(t|\mathbf{z})$,
respectively}%
\label{f:control_surv_test}%
\end{center}
\end{figure}

The illustration of neural networks predicting $K(\mathbf{z},\mathbf{x}_{i})$
and $K(\mathbf{z},\mathbf{y}_{i})$ for a new vector $\mathbf{z}$ of feature
are shown in Fig. \ref{f:control_surv_test}. It can be seen from Fig.
\ref{f:control_surv_test} that the first neural network consists of $c$
subnetworks such that pairs of vectors $(\mathbf{z},\mathbf{x}_{i})$,
$i=1,...,c$, are fed to the subnetworks where $\mathbf{x}_{i}$ is taken from
the dataset of controls. Predictions of the first neural network are $c$
kernels $K(\mathbf{z},\mathbf{x}_{i})$ which are used to compute $\tilde
{S}_{0}(t|\mathbf{z})$ by means of the Beran estimator (\ref{Beran_est}). The
same architecture has the neural network for predicting kernels $K(\mathbf{z}%
,\mathbf{y}_{i})$ used for estimating the treatment SF $\tilde{S}%
_{1}(t|\mathbf{z})$. This network consists of $s$ subnetworks and uses vectors
$\mathbf{y}_{i}$ from the dataset of treatments. After computing estimates
$\tilde{S}_{0}(t|\mathbf{z})$ and $\tilde{S}_{1}(t|\mathbf{z})$, we can find
CATE $\tau(\mathbf{z})$.%

Phases of training and computing CATE $\tau(\mathbf{x})$ by means of neural
kernels are schematically shown as Algorithms \ref{alg:Univers_Train} and
\ref{alg:Univers_Test}, respectively.

\begin{algorithm}
\caption{The algorithm for training neural kernels} \label{alg:Univers_Train}

\begin{algorithmic}
[1]\REQUIRE Datasets $\mathcal{C}$ of $c$ controls and $\mathcal{T}$ of $s$
treatments, number $N$ of generated subsets $\mathcal{C}_{i}^{(r)}$ of
$\mathcal{C}$, number of examples in generated subsets $n$

\ENSURE Neural kernels $K(\cdot,\cdot)$ for their use in the Beran estimator
for control and treatment data

\FOR{$i=1$, $i\leq c$ }\FOR{$r=1$, $r\leq N$ }

\STATE Generate subset $\mathcal{C}_{i}^{(r)}\subset\mathcal{C}\backslash
(\mathbf{x}_{i},y_{i})$

\STATE Form example $\mathbf{a}_{i}^{(r)}=\left(  \mathcal{C}_{i}%
^{(r)},\mathbf{x}_{i},\delta_{i},f_{i}\right)  $

\ENDFOR\ENDFOR

\STATE Train the weight sharing neural network with the loss function given in
(\ref{BENK_40}) on the set of examples $\mathbf{a}_{i}^{(r)}$
\end{algorithmic}
\end{algorithm}

\begin{algorithm}
\caption{The algorithm for computing CATE for a new feature vector $\bf{z}$}
\label{alg:Univers_Test}

\begin{algorithmic}
[1]\REQUIRE Trained neural kernels, datasets $\mathcal{C}$ and $\mathcal{T}$,
testing example $\mathbf{z}$

\ENSURE CATE $\tau(\mathbf{x})$

\FOR{$i=1$, $i\leq c$ }

\STATE Form pair $(\mathbf{z},\mathbf{x}_{i})$ of vectors by using the dataset
$\mathcal{C}$ of controls

\STATE Feed pair $(\mathbf{z},\mathbf{x}_{i})$ to the trained neural kernel
and predict $K(\mathbf{z},\mathbf{x}_{i})$

\ENDFOR

\FOR{$i=1$, $i\leq s$ }

\STATE Form pair $(\mathbf{z},\mathbf{y}_{i})$ of vectors by using the dataset
$\mathcal{T}$ of treatments

\STATE Feed pair $(\mathbf{z},\mathbf{y}_{i})$ to the trained neural kernel
and predict $K(\mathbf{z},\mathbf{y}_{i})$

\ENDFOR

\STATE Compute $W(\mathbf{z},\mathbf{x}_{i})$, $i=1,...,c$, $W(\mathbf{z}%
,\mathbf{y}_{i})$, $i=1,...,s$

\STATE Estimate $\tilde{S}_{0}(t|\mathbf{x}_{k})$ and $\tilde{S}%
_{1}(t|\mathbf{y}_{k})$ using (\ref{Beran_est})

\STATE Compute $\tau(\mathbf{x})$ using (\ref{BENK_41})
\end{algorithmic}
\end{algorithm}

\section{Numerical experiments}

Numerical experiments for studying BENK and its comparison with available
models are performed by using simulated datasets because the true CATEs are
unknown due to the fundamental problem of the causal inference for real data
\cite{Kunzel-etal-2018a}. This implies that control and treatment datasets are
randomly generated in accordance with predefined outcome functions. Nine models

\subsection{CATE estimators for comparison and their parameters}

For investigating BENK and its comparison, we use nine models which can be
united in three groups (the T-learner, the S-learner, the X-learner) such that
every group is based on three base models for estimating SFs (the RSF, the Cox
model, the Nadaraya-Watson kernel regression and the Beran estimator). The
models are given below in terms of survival models.

\begin{enumerate}
\item The T-learner \cite{Kunzel-etal-2018} is a model which estimates the
control SF $S_{0}(t|\mathbf{z})$ and the treatment SF $S_{1}(t|\mathbf{z})$
for every $\mathbf{z}$. The CATE in this case is defined in accordance with
(\ref{CATE_20}).

\item The S-learner \cite{Kunzel-etal-2018} is a model which estimates SF
$S(t|\mathbf{z},T)$ instead of $S_{0}(t|\mathbf{z})$ and $S_{1}(t|\mathbf{z})$
where the treatment assignment indicator $T_{i}\in\{0,1\}$ is included as an
additional feature to the feature vector $\mathbf{z}_{i}$. As a result, we
have a modified dataset
\begin{equation}
\mathcal{D}=\{(\mathbf{z}_{1}^{\ast},\delta_{1},f_{1}),...,(\mathbf{z}%
_{c}^{\ast},\delta_{c},f_{c}),(\mathbf{z}_{c+1}^{\ast},\gamma_{1}%
,h_{1}),...,(\mathbf{z}_{c+s}^{\ast},\gamma_{s},h_{s})\},
\end{equation}
where $\mathbf{z}_{i}^{\ast}=(\mathbf{x}_{i},T_{i})\in\mathbb{R}^{d+1}$ if
$T_{i}=0$, $i=1,...,c$, and $\mathbf{z}_{c+i}^{\ast}=(\mathbf{y}_{i},T_{i}%
)\in\mathbb{R}^{d+1}$ if $T_{i}=1$, $i=1,...,t$. The CATE is determined as
\begin{equation}
\tau(\mathbf{z})=\sum_{j=1}^{s}(t_{j}^{(1)}-t_{j-1}^{(1)})\tilde{S}%
^{(j)}(\mathbf{z},1)-\sum_{k=1}^{c}(t_{k}^{(0)}-t_{k-1}^{(0)})\tilde{S}%
^{(k)}(\mathbf{z},0).
\end{equation}

\item The X-learner \cite{Kunzel-etal-2018} is based on computing the
so-called imputed treatment effects and is represented in the following three
steps. The outcome functions $g_{0}(\mathbf{x})$ and $g_{1}(\mathbf{y})$ are
estimated using a regression algorithm. Second, the imputed treatment effects
are computed as follows:
\begin{equation}
D_{1}(\mathbf{y}_{i})=h_{i}-g_{0}(\mathbf{y}_{i}),\ \ D_{0}(\mathbf{x}%
_{i})=g_{1}(\mathbf{x}_{i})-f_{i}.
\end{equation}

Third, two regression functions $\tau_{1}(\mathbf{y})$ and $\tau
_{0}(\mathbf{x})$ are estimated for imputed treatment effects $D_{1}%
(\mathbf{y})$ and $D_{0}(\mathbf{x})$, respectively. CATE for a point
$\mathbf{z}$\ is defined as a weighted linear combination of the functions
$\tau_{1}(\mathbf{z})$ and $\tau_{0}(\mathbf{z})$ as $\tau(\mathbf{z}%
)=\alpha\tau_{0}(\mathbf{z})+(1-\alpha)\tau_{1}(\mathbf{z})$, where $\alpha
\in\lbrack0,1]$ is a weight which is equal to the ratio of treated patients
\cite{Kunzel-etal-2018a}. The original X-learner does not deal with censored
data. Therefore, we propose a simple survival modification of the X-learner.
It is assumed that $g_{0}(\mathbf{y}_{i})$ and $g_{1}(\mathbf{x}_{i})$ are
expectations $E_{0}(\mathbf{y}_{i})$ and $E_{1}(\mathbf{x}_{i})$ of times to
event corresponding to control and treatment data, respectively. Expectations
$E_{0}(\mathbf{y}_{i})$ and $E_{1}(\mathbf{x}_{i})$ are computed by means of
one of the algorithms for determining estimates of SFs $S_{0}(t|\mathbf{z})$
and $S_{1}(t|\mathbf{z})$.
\end{enumerate}

Estimation of SFs $S_{0}(t|\mathbf{z})$ and $S_{1}(t|\mathbf{z})$ as well as
$S(t|\mathbf{z},T)$ is carried out by means of using the following survival
regression algorithms:

\begin{enumerate}
\item The RSF \cite{Ishwaran-Kogalur-2007}. It is used as the base regressor
to implement other models due to two main reasons. Parameters of random
forests used in experiments are the following:

\begin{itemize}
\item numbers of trees are 50, 100, 200;

\item depths are 2, 4, 6;

\item the smallest values of examples which fall in a leaf are 1 example,
10\%, 20\% of the training set.
\end{itemize}

The above values for the hyperparameters are tested, choosing those leading to
the best results.

\item The Cox proportional hazards model \cite{Cox-1972}. It is used with the
regularization term with the coefficient taking values $0.1$, $0.5$, $1$, $2$,
$5$.

\item The Nadaraya-Watson kernel regression \cite{Nadaraya-1964,Watson-1964}
using the standard Gaussian kernel as the base regression model. In fact, this
regression determines SFs by means of the Beran estimator. In contrast to the
proposed BENK model, we use the standard Gaussian kernel for determining the
attention weights. Values $10^{i}$, $i=-3,...,3$, and also values $0.5$, $5$,
$50$, $200$, $500$, $700$ of the bandwidth parameter of the Gaussian kernel
are tested, choosing those leading to the best results.
\end{enumerate}

In sum, we have $9$ models for comparison whose notations are given in Table
\ref{t:CATE_models}.%

\begin{table}[tbp] \centering
\caption{Notations of the models depending on meta-learners and base models}%
\begin{tabular}
[c]{cccc}\hline
& \multicolumn{3}{c}{Meta-model}\\\hline
Survival regression algorithms & T-learner & S-learner & X-learner\\\hline
Nadaraya-Watson regression & \textbf{T-NW} & \textbf{S-NW} & \textbf{X-NW}%
\\\hline
Cox model & \textbf{T-Cox} & \textbf{S-Cox} & \textbf{X-Cox}\\\hline
RSF & \textbf{T-SF} & \textbf{S-SF} & \textbf{X-SF}\\\hline
\end{tabular}
\label{t:CATE_models}%
\end{table}%

\subsection{Generating synthetic datasets}

We consider two ways for generating random survival times used in numerical
experiments. The first way of generating is the same as it has been done in
\cite{Konstantinov-etal-22}. All numerical experiments are represented on the
basis of simulated datasets such that vectors of features are generated by
means of three functions: the spiral function, the logarithmic function, the
power function. The idea to use these functions stems from the goal to get
complex structures of data, which are poorly processed by many standard
methods. The above functions are defined through a parameter $t$ as follows:

\begin{enumerate}
\item \textbf{Spiral functions:} Two feature vectors having dimensionality $d$
and located on the Archimedean spirals are defined for even $d$ as
\begin{equation}
\mathbf{x}=(t\sin(t),t\cos(t),...,t\sin(t\cdot d/2),t\cos(t\cdot d/2)),
\end{equation}

and for odd $d$ as
\begin{equation}
\mathbf{x}=(t\sin(t),t\cos(t),...,t\sin(t\cdot\left\lceil d/2\right\rceil )).
\end{equation}
Parameter $t$ is uniformly generated from interval $[0,10]$ for controls as
well as for treatments.

\item \textbf{Logarithmic functions:} The feature vectors are logarithms of
parameter $t$ represented as follows:%
\begin{equation}
\mathbf{x}=(a_{1}\ln(t),a_{2}\ln(t),...,a_{d}\ln(t)).
\end{equation}
Values of parameters $a_{1},...,a_{d}$ for performing numerical experiments
with logarithmic functions are uniformly generated from intervals
$[-4,-1]\cup\lbrack1,4]$ for controls as well as for treatments. Values of $t$
are uniformly generated from interval $[0.5,5]$.

\item \textbf{Power functions:} Features are represented as powers of $t$. For
arbitrary $d$ (the number of features), the feature vector is represented as
\begin{equation}
\mathbf{x}=(t^{1/\sqrt{d}},t^{2/\sqrt{d}},...,t^{d/\sqrt{d}}).
\end{equation}
It should be noted that the generating model may lead to almost linear
features with $t$ when $0.8<i/\sqrt{d}<1.6$. To complicate the feature
vectors, we replace the feature vectors with these parameters with the
Gaussian noise having the unit standard deviation and the zero expectation,
i.e., $x_{i}\sim\mathcal{N}(0,1)$. Values of $t$ are uniformly generated from
interval $[0,10]$.
\end{enumerate}

For feature vectors $\mathbf{x}$ and $\mathbf{y}$ from control and treatment
group, times to events $f$ and $h$ are generated are generated by using a
modification of the Cox model generator \cite{Bender-etal-2005} as follows:
\begin{equation}
f=-\ln(0.02)/(0.1\cdot\exp(0.5\cdot t)),
\end{equation}%
\begin{equation}
h=-\ln(0.3)/(0.1\cdot\exp(0.15\cdot t)),
\end{equation}
where $t$ has been generated when vectors $\mathbf{x}$ for the spiral,
logarithmic and power functions were determined.

This way for generating $f$ and $g$ is in agreement with the Cox model. Hence,
we can use the Cox model as a base model among RSFs and the Nadaraya-Watson
regression with Gaussian kernels in numerical experiments.

The number of censored data, denoted as $p$, is taken 25\% of all
observations. Hence, parameters $\delta_{i}$ and $\gamma_{i}$ are generated
from the binomial distribution with probabilities $\Pr\{\delta_{i}%
=1\}=\Pr\{\gamma_{i}=1\}=0.75$, $\Pr\{\delta_{i}=0\}=\Pr\{\gamma_{i}=0\}=0.25$.

The root mean squared error (RMSE) as a measure of the regression model accuracy is
used. The RMSE is computed by using $1000$ randomly selected feature vectors
$\mathbf{z}$. The proportion of treatments and controls in most experiments is
$20\%$ except for experiments investigating how the proportion of treatments
impacts on the RMSE, where the proportion of treatments and controls is denoted
as $q$. For example, if $100$ controls are generated for an experiment with
$q=0.2$, then $20$ treatments are generated in addition to controls such that
the total number of examples is $120$. The generated feature vectors in all
experiments consist of $10$ features. To select optimal hyperparameters of all
regressors, additional validation examples are generated such that the number
of controls is $50\%$ of the training examples from the control group. The
validation examples are not used for training.

\begin{figure}
[ptb]
\begin{center}
\includegraphics[
height=2.2681in,
width=2.7035in
]%
{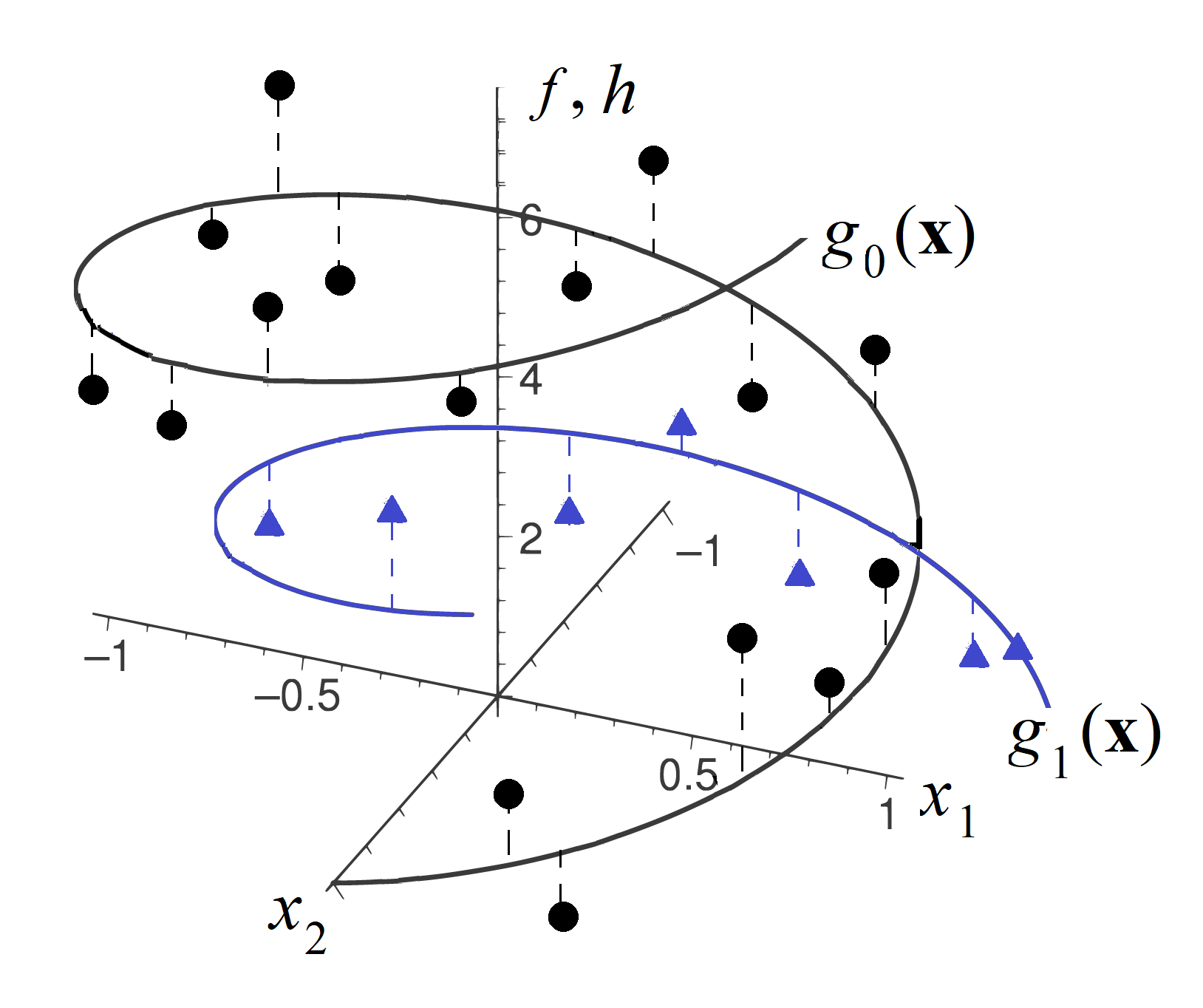}%
\caption{Illustration of the noise added to functions $g_{0}(\mathbf{x})$ and
$g_{1}(\mathbf{x})$}%
\label{f:spirals_eps}%
\end{center}
\end{figure}

Numerical results are obtained under condition that there is some noise
$\varepsilon$ which is added to functions $g_{0}$ and $g_{1}$ or to $f$ and
$h$ as it is depicted in Fig. \ref{f:spirals_eps}. The noise is generated in
accordance with the normal distribution with the zero mean and the standard
deviation $\sigma$ such that $\varepsilon$ is defined as the proportion of
$3\sigma$ and the mean values of $f$ and $h$. Most numerical experiments are
performed under condition $\varepsilon=0.05$. However, a part of experiments
illustrate how the noise impact on the CATE prediction. In these experiments,
different values of $\varepsilon$ are used to study how the noise impacts on
predictions of different models and BENK.%

\subsection{Study of the BENK properties}

In all pictures illustrating results of numerical experiments, dotted curves
correspond to the T-learner (triangle markers), the S-learner (triangle
markers), the X-learner (the circle marker) under condition of using the
Nadaraya-Watson regression with the Gaussian kernel. Dash-and-dot curves
correspond to the Cox models. Dashed curves with the same markers correspond
to the same models implemented by using RSFs. The solid curve with cross
markers corresponds to BENK.

\begin{figure}
[ptb]
\begin{center}
\includegraphics[
height=4.6348in,
width=5.5468in
]%
{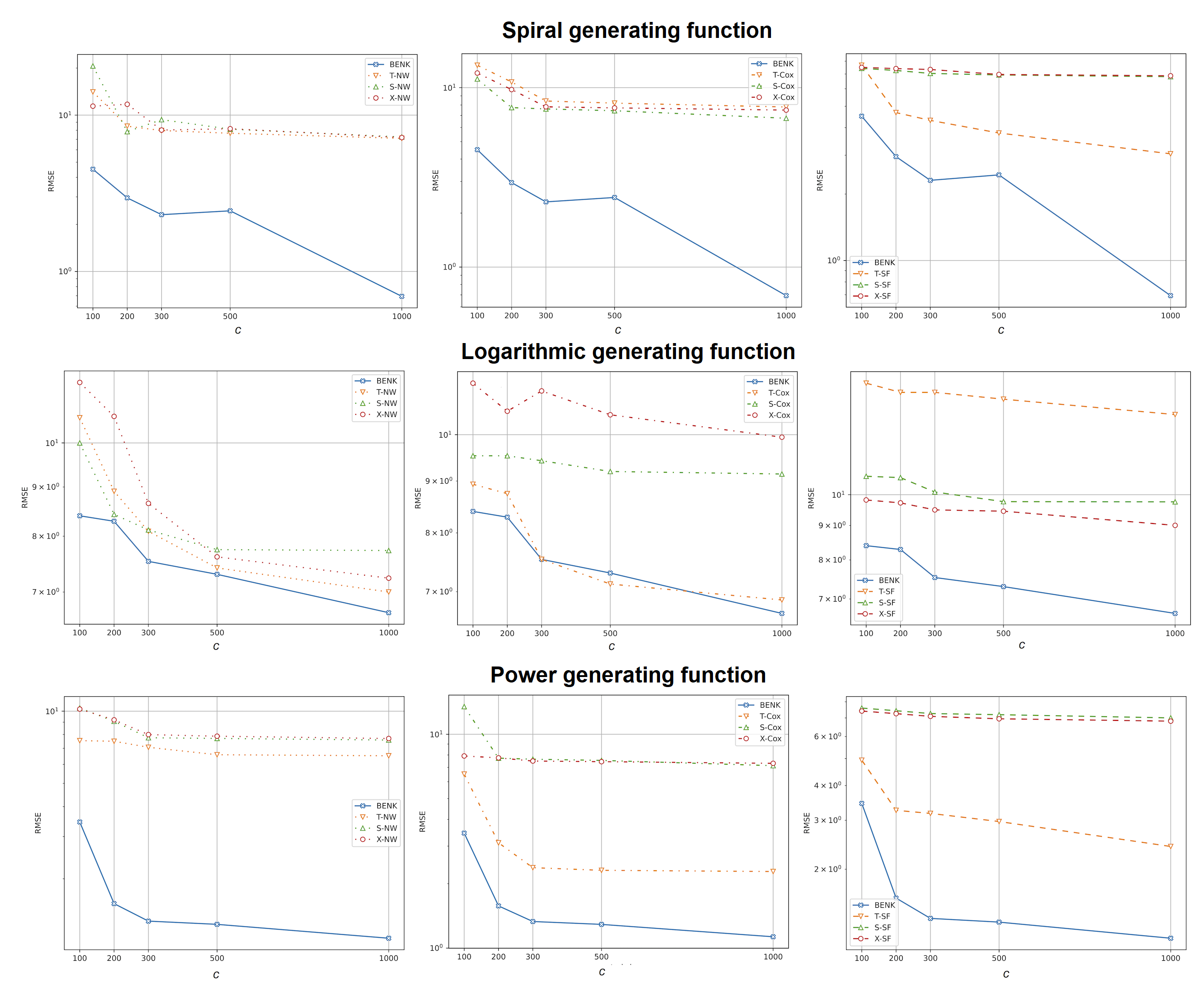}%
\caption{RMSE of the predicted CATE values as a function of the number of
controls when the spiral, logarithmic and power functions are used for
generating instances, and CATE\ is computed by using the Nadaraya-Watson
regression (left pictures), the Cox model (central pictures), the RSF (right
pictures)}%
\label{f:size_all}%
\end{center}
\end{figure}

First, we study different CATE estimators by different numbers $c$ of control
taking values $100$, $200$, $300$, $500$, $1000$. The number of treatments is
determined as $20\%$ of the number of controls. Values of $n$ are $0.2\cdot
c$. Fig. \ref{f:size_all} illustrates how RMSE of the CATE values depends on
the number $c$ of controls for different estimators when different functions
are used for generating examples. Left pictures in Fig. \ref{f:size_all} show
the difference between BENK and models T-NW, S-NW, X-NW. Central pictures show
similar dependencies when models T-Cox, S-Cox, X-Cox are used. Right pictures
illustrate the relationship between CATE values predicted by BENK, T-SF, S-SF,
X-SF by different values $c$. It can be seen from Fig. \ref{f:size_all} that
the proposed model BENK provides better results in comparison with other
models. The largest relative difference between BENK and other models can be
observed when the feature vectors are generated in accordance with the spiral
function. This function produces the most complex data structure such that the
studied models cannot cope with it. On the other hand, it follows from Fig.
\ref{f:size_all}, illustrating the predicted CATE values for the case of the
logarithmic generating function, that models T-NW, S-NW, X-NW provide
comparative and sometimes better results than BENK. This is due to usage of
the Gaussian kernels which better adjust to the logarithmic locations of
feature vectors. The same can be said about models T-Cox, S-Cox, X-Cox based
on the Cox model.%

\begin{figure}
[ptb]
\begin{center}
\includegraphics[
height=3.1151in,
width=5.6928in
]%
{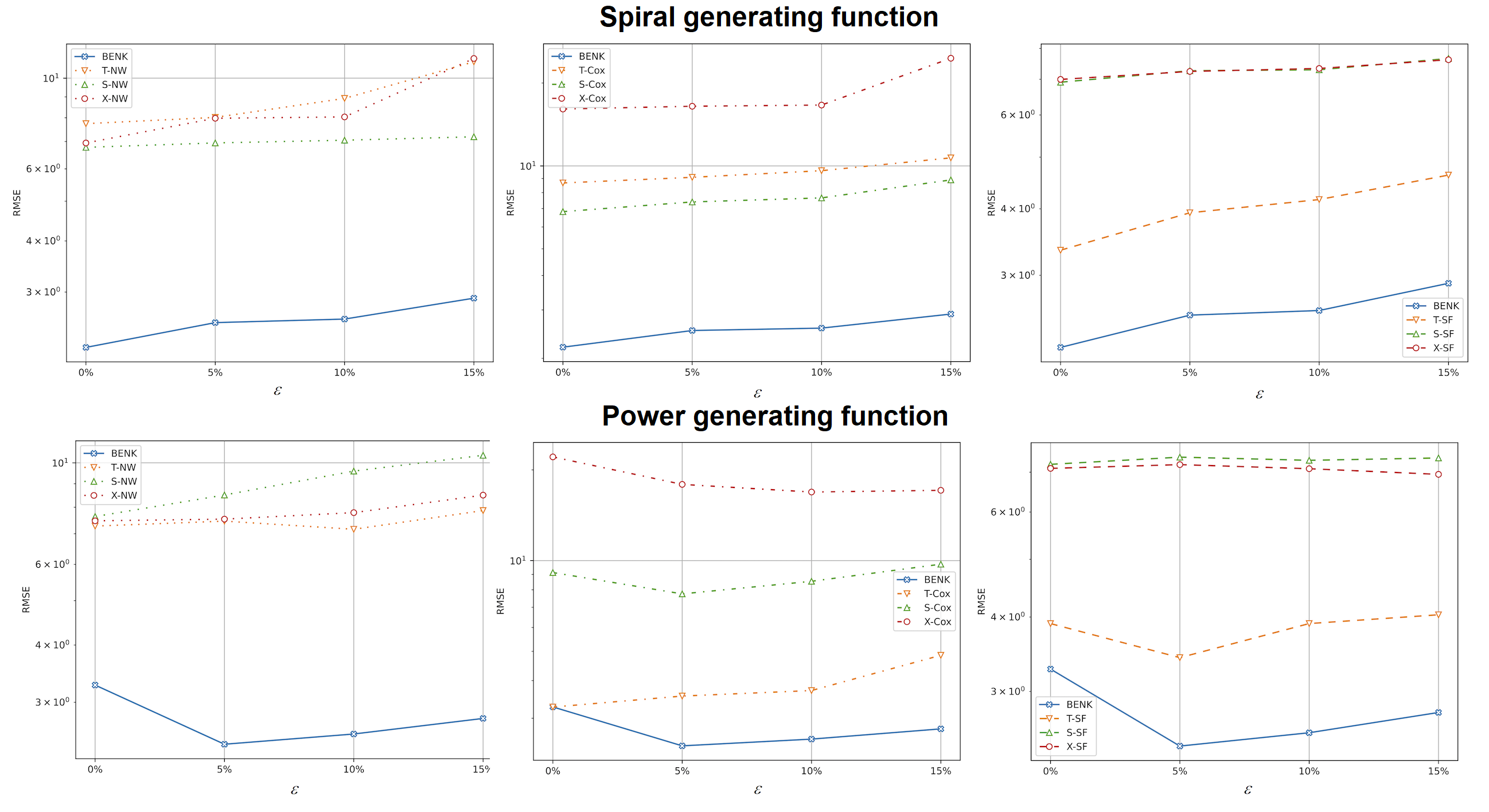}%
\caption{RMSE of the predicted CATE values as functions of the noise value
$\varepsilon$ of controls and treatments when the spiral and power functions
are used for generating examples for models: BENK, T-NW, S-NW, X-NW (the left
picture) and BENK, T-Cox, S-Cox, X-Cox (the central picture) and BENK, T-SF,
S-Cox, X-Cox (the right picture)}%
\label{f:noise_spir_pow}%
\end{center}
\end{figure}

It is important to remind that the above numerical results are obtained under
condition that noise $\varepsilon$ is 5\%. Therefore, the next question is how
the CATE estimators depend on the parameter $\varepsilon$. Numerical results
illustrating how the noise impact on the CATE estimators for the spiral and
power generating functions are shown in Fig. \ref{f:noise_spir_pow}. It can be
seen from Fig. \ref{f:noise_spir_pow} that the relative difference between the
model RMSE measures obtained for the spiral generating function is not change
with increase of the noise. However, one can see that this difference
increases then the power function is used for generating the feature vectors.
In all cases, we again observe outperforming results provided by BENK.%

\begin{figure}
[ptb]
\begin{center}
\includegraphics[
height=1.6842in,
width=5.7368in
]%
{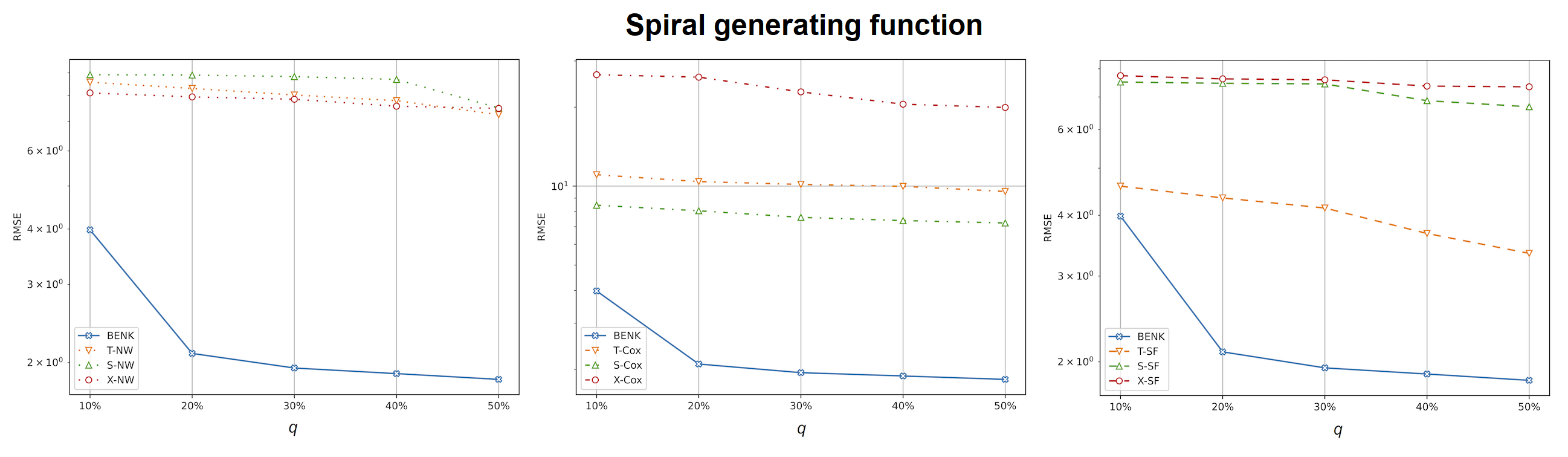}%
\caption{RMSE of the predicted CATE values as functions of on the proportion
$q$ of treatments and controls when the spiral function is used for generating
examples for models: BENK, T-NW, S-NW, X-NW (the left picture) and BENK,
T-Cox, S-Cox, X-Cox (the central picture) and BENK, T-SF, S-Cox, X-Cox (the
right picture)}%
\label{f:parts_treatm_spiral}%
\end{center}
\end{figure}

Another interesting question is how the CATE estimators depend on the
proportion $q$ of treatments and controls in the training set. The
corresponding numerical results are shown in Fig. \ref{f:parts_treatm_spiral}
for the spiral generating function. One can see from Fig.
\ref{f:parts_treatm_spiral} that improvement of the RMSE is sufficient in
comparison with other CATE estimators when $q$ is changed from 10\% to 20\%.
Moreover, we again observe the outperformance of BENK in comparison with other
estimators. It should be noted that similar results take place when other
generating functions (logarithmic and power) are used.%

\begin{figure}
[ptb]
\begin{center}
\includegraphics[
height=1.6543in,
width=5.7719in
]%
{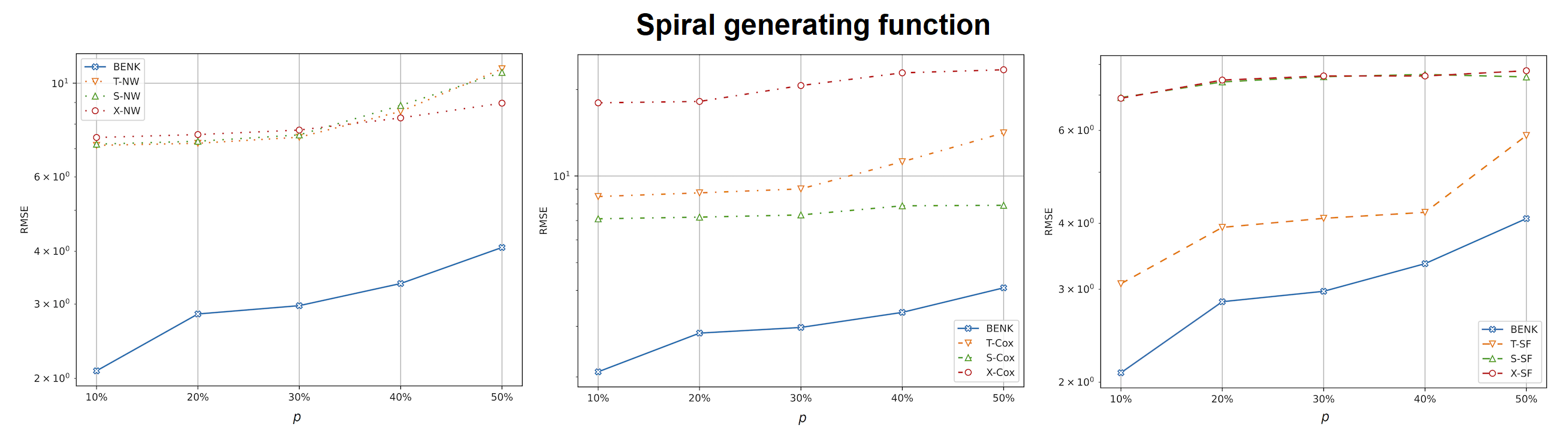}%
\caption{RMSE of the predicted CATE values as functions of on the ratio $p$ of
censored data when the spiral function is used for generating examples for
models: BENK, T-NW, S-NW, X-NW (the left picture) and BENK, T-Cox, S-Cox,
X-Cox (the central picture) and BENK, T-SF, S-Cox, X-Cox (the right picture)}%
\label{f:cens_spiral}%
\end{center}
\end{figure}

In the previous experiments, the amount of censored data was taken $p=25\%$ of
all observations. However, it is interesting to study how this amount impacts
on the RMSE of the CATE estimators. Fig. \ref{f:cens_spiral} illustrates the
corresponding dependences when the spiral generating function is used. It can
be seen from Fig. \ref{f:cens_spiral} that RMSE measures of all estimators,
including BENK, increase with the amount of censored data.%

\qquad

Figs. \ref{f:size_all}-\ref{f:cens_spiral} can be viewed as qualitative
results of comparing different estimators. Table \ref{t:survhte_1} aims to
quantitatively compare results under the following conditions: $c=200$,
$s=60$, $p=0.3$, $m=20$, $N=1000$, $\varepsilon=0.1$. One can see from Table
\ref{t:survhte_1} that BENK provides ouperforming results.%

\begin{table}[tbp] \centering
\caption{The RMSE values of CATE for different models by different generating functions}%
\begin{tabular}
[c]{cccc}\hline
& \multicolumn{3}{c}{Generating functions}\\\hline
Model & Spiral & Logarithmic & Power\\\hline
T-NW & $9.984$ & $7.223$ & $7.361$\\\hline
S-NW & $7.387$ & $8.978$ & $7.565$\\\hline
X-NW & $9.257$ & $9.339$ & $7.199$\\\hline
T-Cox & $12.28$ & $6.824$ & $3.312$\\\hline
S-Cox & $7.221$ & $6.807$ & $7.412$\\\hline
X-Cox & $14.67$ & $9.419$ & $11.91$\\\hline
T-SF & $7.245$ & $7.205$ & $5.113$\\\hline
S-SF & $7.276$ & $7.104$ & $7.094$\\\hline
X-SF & $6.754$ & $9.297$ & $7.875$\\\hline
BENK & $\mathbf{6.215}$ & $\mathbf{5.456}$ & $\mathbf{2.518}$\\\hline
\end{tabular}
\label{t:survhte_1}%
\end{table}%

It should be noted that we did not provide results of various deep neural
network extensions of the CATE estimators because they have not been
successful. The problem is that neural networks require a large amount of data
for training and the considered small datasets have led the networks to
overfitting. That is why we studied models which provide satisfactory
predictions under condition of small data.

\section{Conclusion}

BENK as a new method for solving the CATE problem under censored data has
been presented. It extends the idea behind TNW-CATE proposed in
\cite{Konstantinov-etal-22} to the case of censored data. In spite of many
similar parts of TNW-CATE and BENK they are different because BENK is based on
using the Beran estimator for training and can be successfully applied to
survival analysis of controls and treatments. However, TNW-CATE and BENK use
the same idea to train neural kernels implemented as neural networks instead
of used standard kernels.

It is also interesting to point out that BENK does not require to have a large
dataset for training though the neural network is used for implementing the
kernels. This is due to a special way which is proposed to train the network
and considers pairs of examples from the control group for training like the
Siamese neural networks. Our experiments have illustrated the outperforming
characteristics of BENK. At the same time, we have to point out some
disadvantages of BENK. First, it has many tuning parameters, including
parameters of the neural network, parameters of training $n$ and $N$, such
that the training time may be sufficiently increased in comparison with other
methods of solving the CATE problem. Second, BENK assumes that the feature
vector domains are similar for controls and treatments. It does not mean that
they have to totally coincide, but the corresponding difference of domains
should not be very large. A method which could take into account a possible
difference between the feature vector domains for controls and treatments can
be regarded as a direction for further research. An idea behind the method is
to combine the domain adaptation models and BENK.

Another direction for further research is to study robust versions of BENK
when there are anomalous observations which may impact on training the neural
network. An idea behind the robust version is to use attention weights for
feature vectors, but also to introduce additional attention weights for predictions.

It should be noted that the Beran estimator is one of several estimators which
are used in survival analysis. Moreover, we have studied only the difference
in expected lifetimes as a definition of CATE in the case of censored data.
There are other definitions, for instance, the difference in SFs and the
hazard ratio, which may lead to more interesting models. Therefore, the BENK
implementations and studies by using other estimators and definitions of CATE
can be also considered as directions for further research.

\bibliographystyle{unsrt}
\bibliography{Attention,Boosting,Deep_Forest,Explain,MYBIB,MYUSE,Survival_analysis,Transf_Learn,Treatment}

\end{document}